\documentclass[letterpaper, 10 pt, conference]{ieeeconf}  

\usepackage{graphicx}
\usepackage{cite}
\usepackage{amssymb}
\usepackage{mathrsfs}
\usepackage{calligra}
\usepackage{amsmath}
\usepackage{multirow}
\usepackage{float}
\usepackage{hyperref} 

\IEEEoverridecommandlockouts                              

\title{\LARGE \bf
CMIP-CIL: A Cross-Modal Benchmark for\\Image-Point Class Incremental Learning}

\author{Chao Qi$^{1}$, Jianqin Yin$^{1,*}$, and Ren Zhang$^{1}$
\thanks{$^{1}$School of Intelligent Engineering and Automation, Beijing University of Posts and Telecommunications, Beijing 102206, China.
        }%
        \thanks{*Corresponding author}
        \thanks{e-mail: qichao199@163.com}
}

\begin{document}

\maketitle
\thispagestyle{empty}
\pagestyle{empty}

\begin{abstract}

Image-point class incremental learning helps the 3D-points-vision robots continually learn category knowledge from 2D images, improving their perceptual capability in dynamic environments. However, some incremental learning methods address unimodal forgetting but fail in cross-modal cases, while others handle modal differences within training/testing datasets but assume no modal gaps between them. We first explore this cross-modal task, proposing a benchmark CMIP-CIL and relieving the cross-modal catastrophic forgetting problem. It employs masked point clouds and rendered multi-view images within a contrastive learning framework in pre-training, empowering the vision model with the generalizations of image-point correspondence. In the incremental stage, by freezing the backbone and promoting object representations close to their respective prototypes, the model effectively retains and generalizes knowledge across previously seen categories while continuing to learn new ones. We conduct comprehensive experiments on the benchmark datasets. Experiments prove that our method achieves state-of-the-art results, outperforming the baseline methods by a large margin. The code is available at \href{https://github.com/chaoqi7/CMIP-CIL}{https://github.com/chaoqi7/CMIP-CIL}.
\end{abstract}

\section{INTRODUCTION}

Humans continually learn to recognize objects from 2D images in books, putting the knowledge into practice in the 3D real world. The \textit{cross-modal continual learning} ability also greatly benefits intelligent robots, helping them achieve multimodal knowledge with only 2D image training.

\begin{figure}
    \centering
    \includegraphics[width=1\linewidth]{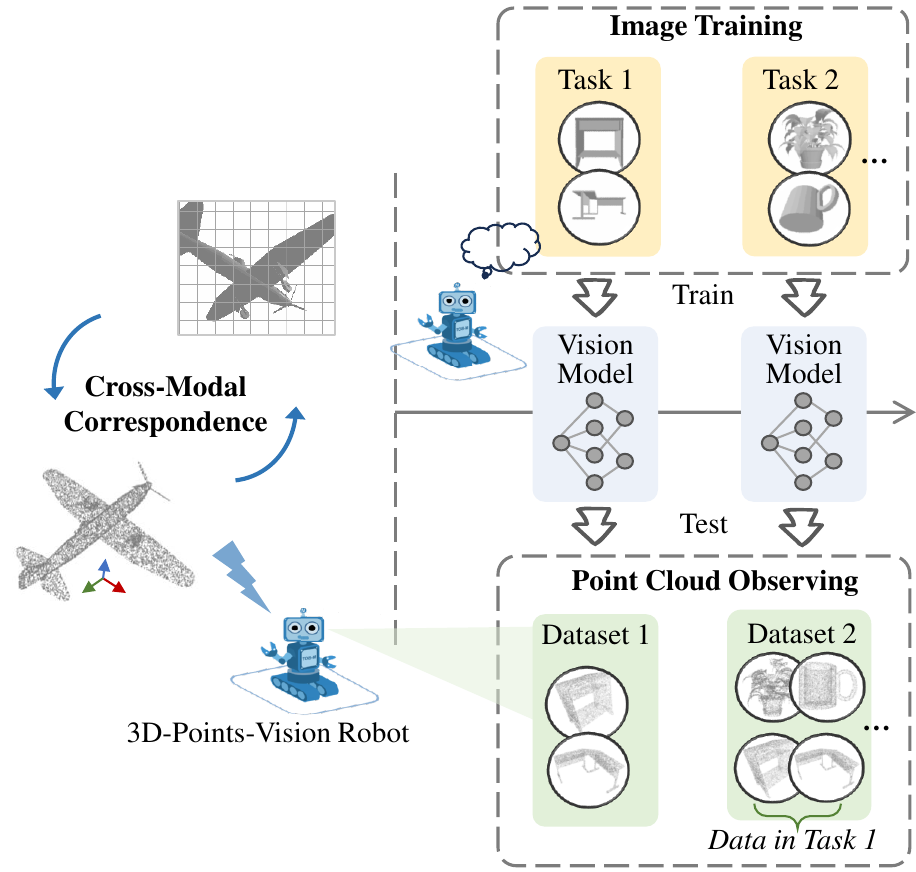}
    \caption{Task of IP-CIL. Task 1: learns to classify objects in images, testing with point-cloud-based classifications. Task 2: learn new objects in images, testing point cloud ones in the current and former classes—the same in the following tasks.}
    \label{fig1}
\end{figure}

This paper explores a branch of continual learning, class incremental learning, proposing a cross-modal benchmark to help the 3D-points-vision robots continually learn category knowledge from 2D images. The \textbf{I}mage-\textbf{P}oint \textbf{C}lass \textbf{I}ncremental \textbf{L}earning (IP-CIL) is illustrated in Fig. \ref{fig1}. It empowers robots to adapt to evolving tasks efficiently in dynamic environments, improving their perceptual capability in real-world applications.

General Class Incremental Learning (CIL) focuses on relieving the model's catastrophic forgetting of knowledge learned in prior tasks. The methods can be divided into the following categories \cite{RN449} data replay \cite{RN450, RN429,RN451}, knowledge distillation \cite{RN441, RN442}, regularization \cite{RN436, RN430}, rectification \cite{RN446, RN445}, and dynamic network methods \cite{RN431, RN417, RN432}. As a variant of the dynamic network, the pre-trained model method freezes the backbones to remember former knowledge and introduces trainable layers to adapt novel objects. It has achieved state-of-the-art CIL results recently. However, these methods focus on the unimodal forgetting issues, which cannot be used directly on cross-modal data.

Even though IP-CIL has never been explored before, CIL issues crossing other modalities \cite{RN474, RN475, RN488, RN489} have recently received attention, in which vision-language ones are the most widely discussed. \cite{RN474} pioneered the vision-language CIL, followed by a series of studies exploring different cross-modal methods \cite{RN482, RN484, RN476, RN477}. There are modal differences within these studies' training/testing datasets, but no modal gaps exist between the training and testing datasets. Differently, the modal of training data (pure images) differs from that of the testing point clouds in our settings. 

\textbf{In summary, the biggest challenge of the IP-CIL task is to relieve cross-modal catastrophic forgetting.} To address this challenge, we should empower the vision model to build the image-point correspondence of objects, and it should be generalizable along with the incremental classes. If the model can remember the image-based object in the incremental stage, it can also recognize that in point clouds. Besides, we should also explore an efficient method to relieve the model's forgetting of former category knowledge. Thus, the vision model can continually learn new knowledge while not forgetting old ones.

Specifically, we propose a cross-modal \textbf{CIL} benchmark embedded with \textbf{C}ontrastive \textbf{M}asked \textbf{I}mage-\textbf{P}oint pre-training (\textbf{CMIP-CIL}). It builds the image-point correspondence of each object by contrastive learning in pre-training. The traditional contrastive method easily falls into the domain shift trap, losing generalizability while observing objects never seen before. To address this problem, we randomly mask the point clouds and generate multi-view images with differentiable renderers. The numerous image-point pairs extend the perception domains of the contrastive models. Thus, the model can be generalizable along with the incremental classes.

In the continual learning stage, the backbone network is frozen to maintain stable foundational representations, allowing the model to revisit image-point correspondence learned during pre-training. Trainable layers, combined with a regularization function, encourage representations of image-based objects within the same category to become more similar, reducing intra-class variability and consolidating class-specific features. By promoting object representations close to their respective prototypes, the model effectively retains and generalizes knowledge across previously seen categories while continuing to learn new ones. Our contributions can be summarized as:

\begin{itemize}
    \item We propose a cross-modal benchmark CMIP-CIL. It is the first study that helps the vision models continually recognize objects in point clouds with category knowledge learned from 2D images.
    \item We propose a contrastive learning method based on masked image-point pairs. It helps the model build the image-point correspondence of objects and is generalizable along with the incremental classes.
    \item Ours achieves SOTA results on different datasets with different class incremental settings, outperforming the baseline methods by a large margin.
\end{itemize}

\begin{figure*}[h]
    \centering
    \includegraphics[width=0.90\linewidth]{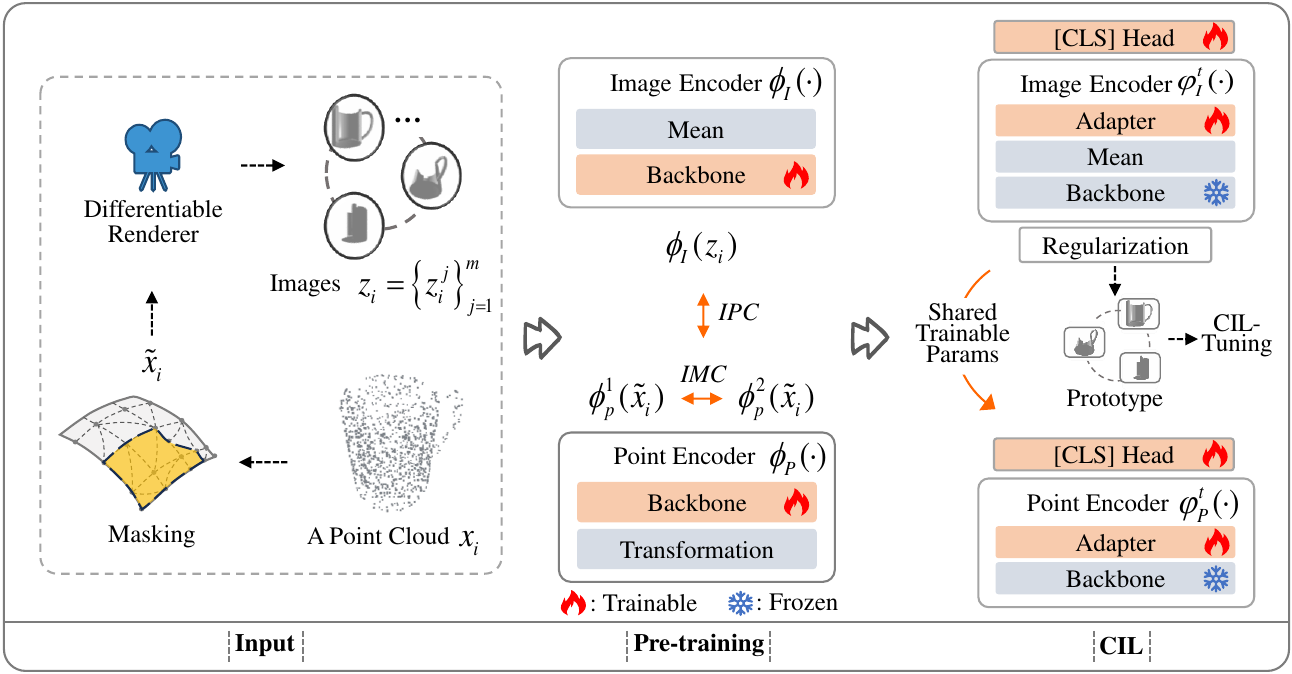}
    \caption{Framework of the CMIP-CIL benchmark. Through image rendering with random masking points, image-point pairs $\{ z_i^j\} _{j = 1}^m \sim {\tilde x_i}$ are generated. Image-point contrastive (IPC) and intra-modal contrastive (IMC) narrow the gap between image encoding ${\phi _I}({z_i})$ and point encoding $\{ \phi _p^1({\tilde x_i}),\phi _p^2({\tilde x_i})\}$ for the same object. In CIL, novel encoders (with trainable layers) $\varphi _I^t( \cdot )$, $\varphi _P^t( \cdot )$ cooperate with the regularization item to tune the class prototypes in task \textit{t}.}
    \label{fig2}
\end{figure*}

\section{Related Work}

\subsection{Class Incremental Learning}

Various methods are explored in CIL \cite{RN450, RN429, RN486}. The data replay ones \cite{RN450, RN429, RN451} use \textit{exemplars} (samples in former classes) to remember the prior learning knowledge. This category of method cooperates well with others and is widely used. The distillation \cite{RN441, RN442} methods use the teach-student manner to distill knowledge while exemplar training, reducing the model's catastrophic forgetting in the continual learning stage. Besides, some methods minimize prior knowledge forgetting through parameter regularization \cite{RN436, RN430} and biased prediction rectification \cite{RN446, RN445, RN444}.

Dynamic network methods have received attention recently \cite{RN431, RN417,RN432}, and the pre-training model method \cite{RN448, RN421} is a variant. It leverages a fixed backbone network to retain knowledge from previously learned classes. The pre-training model method introduces trainable layers or modules that can be updated during incremental training to adapt to newly observed objects. This approach effectively balances the preservation of prior knowledge with acquiring new information, thereby reducing catastrophic forgetting. By freezing the backbone network, these methods ensure that the foundational representations of earlier classes remain stable while the additional trainable layers allow the model to adjust to new tasks flexibly.

However, these methods are explored and verified in image-based CIL, which cannot be directly used in the cross-modal domains, including the image-point ones.

\begin{figure*}[h]
    \centering
    \includegraphics[width=0.90\linewidth]{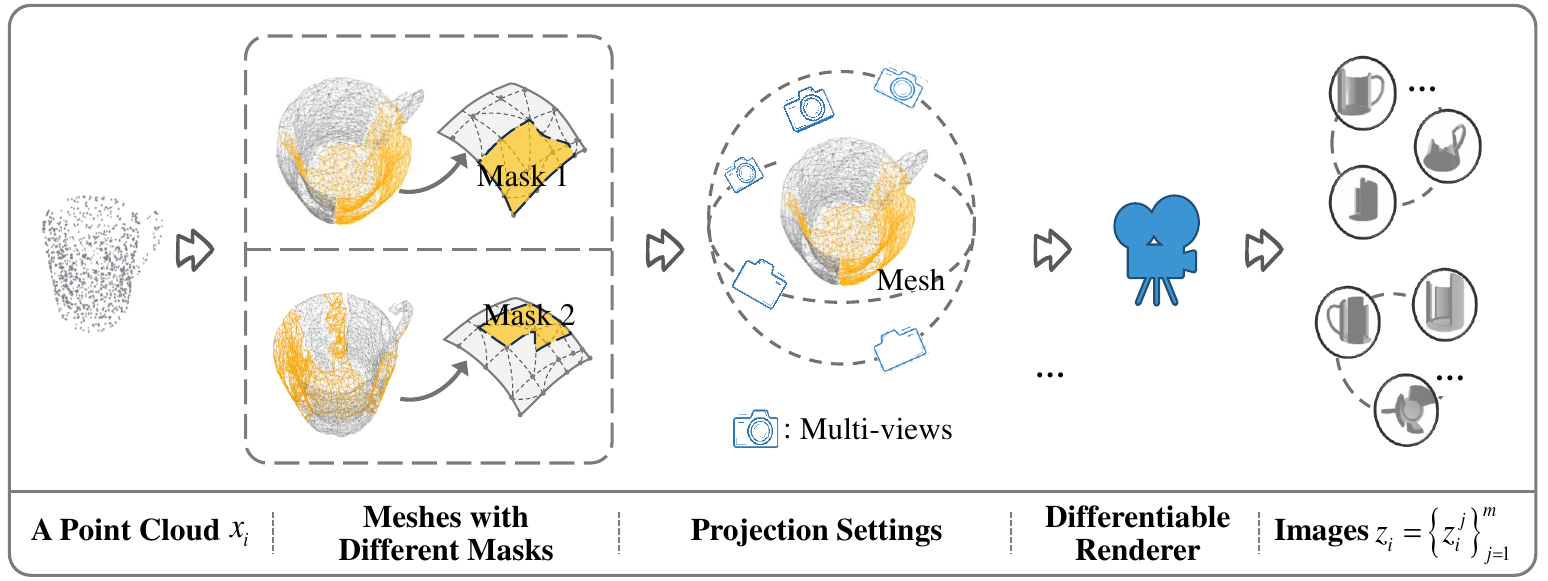}
    \caption{RRM illustration. Considering a point cloud as the input. Randomly masked meshes are projected with a differentiable renderer to generate multi-view images.}
    \label{fig3}
\end{figure*}

\subsection{Cross-modal Class Incremental Learning}

Vision-language \cite{RN474}, vision-audio \cite{RN475}, and vision-sensors (acceleration, gyroscope, etc.)\cite{RN488, RN489} CILs have been discussed recently. \cite{RN474} proposed a benchmark in the vision-language CIL, followed by a series of studies \cite{RN482, RN484, RN476, RN477}. \cite{RN482} introduced a contrastive language-image pre-training (CLIP) model. This study observes that the CLIP's knowledge transfer ability significantly degrades the model's catastrophic forgetting. \cite{RN477} presented a cross-modal alternating learning framework with task-aware representations—that effectively utilizes visual and linguistic information to advance continual learning capabilities. \cite{RN484} and \cite{RN476} proposed self-critical and attention distillation methods, respectively, addressing the forgetting problem by transferring knowledge from previous domains. 

The above methods balance preserving previously learned knowledge and adapting to new data, which inspires us greatly. However, these studies address the domain gap problem inside the training/testing dataset. Differently, our focus is on addressing the domain gap between the training and testing datasets. It motivates us to explore a way to address this image-point domain gap in CIL.

\section{Problem Statement}

The image-point class incremental learning (IP-CIL) can be formulated as follows: a sequence of \textit{T} training datasets $D = \{ {D_1},{D_2}, \cdots ,{D_T}\}$ is given, and ${D_t} = \{ ({z_i},{y_i})\} _{i = 1}^{{n_t}}$ includes ${n_t}$ samples. Every training sample ${z_i} \in {\mathbb{R}^{m \times h \times w \times c}}$ is a \textit{m}(ultiview) image with \textit{h}(eight)×\textit{w}(idth)×\textit{c}(hannel) size, and ${y_i} \in {\mathcal{Y}_t}$ is the image's category label. Each task has a label space ${\mathcal{Y}_t}$, and no task overlaps with others. In the testing data $\tilde D = \{ {\tilde D_1},{\tilde D_2}, \cdots ,{\tilde D_T}\} $, every sample ${\tilde x_i} \in {\mathbb{R}^{m \times c}}$ in ${\tilde D_t}$ is a point cloud with \textit{m} points. 

In task \textit{t}, the model $f( \cdot )$ learns image-based ${D_t}$ with lable space ${\mathcal{Y}_t}$ and predicts on point-based ${\tilde D_t}$. However, trainable parameters keep updating, causing the model to forget prior knowledge from ${D_1} \cup  \cdots {D_{t - 1}}$
 with label spaces ${Y_1} \cup  \cdots {Y_{t - 1}}$. It is called catastrophic forgetting. We aim to train a model with cross-modal prediction ability and relieve the model of forgetting knowledge.

\section{Method}
\subsection{Overview}

Fig. \ref{fig2}. illustrates the framework of the CMIP-CIL benchmark, a novel approach for continual learning across modalities. The process begins by generating massive image-point pairs by randomly masking 3D point clouds and rendering corresponding 2D images. This enables the model to develop a generalizable ability to establish image-point correspondence during pre-training. This alignment of 2D visual features with 3D spatial structures builds a robust foundation for incremental learning.

In CIL phase, shared trainable layers adapt to novel object categories, and the backbone is frozen, dynamically adjusting to new inputs while preserving prior knowledge. The framework also refines the class prototype—a representative feature vector for each category—by continuously tuning it based on incoming data, enhancing the model's ability to distinguish between classes and maintain high accuracy as new categories are introduced.

\subsection{Image Rendering with Randomly Masking Points}

Even though contrastive learning has been widely used across different modalities \cite{HuWZCG24}, limited data with specific semantic labels makes the model fall into the domain shift trap easily. A lack of generalization ability harms the CIL, in which novel objects are seen continually. 

To enhance the model's generalization ability, we feed it diverse data with semantic meaninglessness by deliberately removing object surfaces, which blurs their semantic information. Specifically, we convert a point cloud ${x_i}$ into a mesh and randomly mask continuous faces, generating numerous semantic-agnostic meshes that still retain rich geometric information. This process ensures the model focuses on structural patterns rather than relying solely on semantic cues. To further enrich the training data, we configure different projection settings for the differentiable renderer \cite{RN416}, producing \textit{m}(ultiview) images ${z_i} = \left\{ {z_i^j} \right\}_{j = 1}^m$ that capture the object's geometry from various perspectives. This approach not only introduces variability into the training process but also strengthens the model's ability to generalize across different modalities. The process of image \textbf{R}endering with \textbf{R}andomly \textbf{M}asking points (\textbf{RRM}) is shown in Fig. \ref{fig3}.

\subsection{Image-Point Contrastive Learning}

Given the masked image-point pairs $\{ z_i^j\} _{j = 1}^m \sim {\tilde x_i}$ in the pre-training stage, we introduce contrastive learning \cite{RN567, RN568} to train the vision model to establish generalizable image-point correspondence for objects (Fig. \ref{fig2}). Contrastive learning is designed to learn meaningful representations by maximizing the similarity between positive embedding pairs while minimizing it for negative pairs. We employ the NT-Xent loss \cite{RN569} (Normalized Temperature-Scaled Cross Entropy Loss) to guide the contrastive learning process:
\begin{multline}
    l(e_i^1,e_i^2) = \\ -\log  \frac{\exp (s(e_i^1,e_i^2)/\tau )}
    {\sum\limits_{k = 1,k \ne i}^N \exp (s(e_i^1,e_k^1)/\tau ) 
    + \sum\limits_{k = 1}^N \exp (s(e_i^1,e_k^2)/\tau )}
\label{eq1}
\end{multline}
Where $s( \cdot , \cdot )$  indicates the cosine similarity function, $\tau$ is the temperature coefficient, $e_i^1$ and $e_i^2$ are the embedding pairs. 

We introduce ${L_{IMC}}$ to enhance \textbf{I}ntra-\textbf{M}odal \textbf{C}onsistency by promoting similarity among point-based objects with varying poses; ${L_{IPC}}$ to strengthen \textbf{I}mage-\textbf{P}oint \textbf{C}onsistency by maximizing similarity between embedding pairs \cite{RN599}. Specifically, ${L_{IMC}}$ is defined as follows:
\begin{multline}
    {\mathcal{L}_{IMC}} = \frac{1}{{2N}}\sum\limits_{i = 1}^N {[l} (\phi _p^1({\tilde x_i}),\phi _p^2({\tilde x_i})) + l(\phi _p^2({\tilde x_i}),\phi _p^1({\tilde x_i}))]
\label{eq2}
\end{multline}
Where \textit{N} is the batch size; ${\phi _P}( \cdot )$ is the point encoder containing a trainable backbone with pose transformation function. ${L_{IPC}}$ is defined as:
\begin{multline}
    {\mathcal{L}_{IPC}} = \\ \frac{1}{{2N}}\sum\limits_{i = 1}^N {[l} (\phi _p^1({\tilde x_i}),{\phi _I}(\{ z_i^j\} _{j = 1}^m)) + l({\phi _I}(\{ z_i^j\} _{j = 1}^m),\phi _p^1({\tilde x_i}))]
\label{eq2}
\end{multline}
where ${\phi _I}( \cdot )$ is the image encoder containing a trainable backbone with a multi-view mean function. ${L_{IPC}}$ works with  ${L_{IMC}}$ to ensure that the model effectively aligns 2D image features with their corresponding 3D point cloud representations, also aligning the intra-modal representations. 

\subsection{Prototype Calculations with Regularizations}

We introduce adapters into the encoders to expand the image-point correspondence knowledge and recognize novel objects (Fig. \ref{fig2}). The backbones are frozen, and the adapters share parameters between the image encoder $\varphi _I^t( \cdot )$ and point encoder $\varphi _P^t( \cdot )$ in task \textit{t}, and the same for the classification heads. Thus, the representations of an object with different modalities can be similar. We calculate the prototype using the mean value of image encodings:
\begin{equation}
    {p_{t,\hat y}} = {{\rm \mathbb{E}}_{({z_i},{y_i}) \sim{\xi _{{y_i} = \hat y}}}}[\varphi _I^t({z_i})]
\end{equation}
where ${\xi _{{y_i} = \hat y}}$ is the exemplar set in task \textit{t} while training on data with label $\hat y$, and ${p_{t,\hat y}}$ is the corresponding class prototype. 

We regularize the representation of training samples, promoting the representations to approach the class prototype. $\mathcal{L}_{({z_i},{y_i})\sim{D_{{y_i} = \hat y}}}^1$ denotes the similarity between the class prototype ${p_{t,\hat y}}$ and object representation $\varphi _I^t({z_i})$. The smaller, the more similar. The regularization is a process of minimizing the following values:
\begin{equation}
\mathcal{L}_{({z_i},{y_i})\sim {D_{{y_i} = \hat y}}}^1 = 1 - \mathbb{E}[\frac{{\varphi _I^t({z_i}) \cdot {p_{t,\hat y}}}}{{\left\| {\varphi _I^t({z_i})} \right\| \cdot \left\| {{p_{t,\hat y}}} \right\|}}]
\end{equation}

In the latter tasks, the prototypes of former classes are recalculated with the exemplar sets, resulting in ongoing prototype tunings along the CIL stage. The regularizations help relieve the model's catastrophic forgetting.

\subsection{Loss Function}

In the pre-training stage, we aim to build the image-point correspondence. The loss ${\mathcal{L}_{pre.}}$ denotes the intra-modal and cross-modal representation gap: ${\mathcal{L}_{pre.}} = {\mathcal{L}_{IMC}} + {\mathcal{L}_{IPC}}$

In the CIL, the cross-entropy item is illustrated as:
\begin{equation}
\mathcal{L}_{({z_i},{y_i})\sim {D_{{y_i} = \hat y}}}^2 = {\rm{CE}}({W^T}\varphi _I^t({z_i}),\hat{y})
\end{equation}

where \textit{W} is the classification head, the total loss in CIL is the combination of a regularization item and a cross-entropy item: ${\mathcal{L}_{CIL}} = {\mathcal{L}^1} + {\mathcal{L}^2}$

\section{Experiments}

We conduct comparison and ablation experiments on benchmark datasets, evaluating the effectiveness of our method in the IP-CIL. Experiments are designed to answer the following questions: (1) \textit{Can our method effectively relieve the model's catastrophic forgetting in the cross-modal continual learning}? \textit{Can our method outperform the baseline methods}? (2) \textit{Does our method perform well in building the image-point correspondence}? (3) \textit{How do several essential designs affect our method's effectiveness}?

Sections \textit{\uppercase\expandafter{\romannumeral 5}}-\textit{A} to \textit{D} introduce the dataset, comparison methods, implementation details, and evaluation metrics in the experiments. Section \textit{\uppercase\expandafter{\romannumeral 5}}-\textit{E} answers question (1) by comparing baselines; section \textit{\uppercase\expandafter{\romannumeral 5}}-\textit{F} shows the visualization results of our method; section \textit{\uppercase\expandafter{\romannumeral 5}}-\textit{G} answers questions (2) and (3) by verifying the image-point correspondence and the effectiveness of several essential designs.

\subsection{Datasets}

A multi-modal vision of the ModelNet40 \cite{RN424} and the ShapeNet55 \cite{RN423} are used as the benchmark datasets, while the image data is for training and point cloud data for testing. ModelNet40 and ShapeNet55 contain 40 and 55 class categories, respectively. These popular point cloud classification datasets are created by collecting 3D CADs from open-source repositories.

We follow the data split setting in \cite{RN414, RN466, RN410}: \textbf{m}ultimodal \textbf{M}odel\textbf{N}et40 with an \textbf{inc}rement of \textbf{4} classes (m-MN40-\textit{Inc}.4) and \textbf{m}ultimodal \textbf{S}hape\textbf{N}et55 with an \textbf{inc}rement of \textbf{6} classes (m-SN55-\textit{Inc}.6; 7 classes in the last stage) are used in the experiment. Besides, m-MN40-\textit{Inc}.8 is introduced as a benchmark.

\subsection{Comparison Methods}

This paper first discusses the IP-CIL; no baselines can be directly used in our experiment. Thus, we introduce the state-of-the-art methods in general CILs and reproduce them on the multimodal datasets for comparisons, verifying our method's superiority. It includes iCaRL \cite{RN416}, WA \cite{RN447}, PODNet \cite{RN441}, and SimpleCIL \cite{RN448}.

\subsection{Implementation Details}

We implement our method with Pytorch and PyCIL \cite{RN592}, a Python toolbox for class-incremental learning, on a single NVIDIA GeForce RTX 4090 and Intel(R) Xeon(R) Gold 6430 CPU.

We project each point cloud in ModelNet40 and ShapeNet55 into 10-view images as the basis multimodal dataset in our experiment. To enhance the generalizations of image-point correspondence, we randomly mask the continuous faces of each mesh-structured point cloud in the initial task, resulting in 20 point clouds with different masks. We conduct image rendering with these masked point clouds, forming a multimodal dataset for pre-training. 

In the pre-training stage, the backbone is trained using back-propagation and SGD optimizer with an initial learning rate of 0.001 and batch size of 16. In CIL, the adapter layers are trained using back-propagation and SGD optimizer with an initial learning rate of 0.01 and batch size of 16.

We follow the exemplar setting in the CIL studies of point clouds \cite{RN414, RN466}, storing a fixed number of samples in the memory for incremental learning. $\mathcal{M}$ (exemplar samples) = 800 while conducting m-MN40-\textit{Inc}.4 and m-MN40-\textit{Inc}.8 experiments, and $ \mathcal{M} \approx 1000$ for m-SN55-\textit{Inc}.6. We follow iCaRL\cite{RN416} to randomly shuffle class orders with seed 1993.

\subsection{Evaluation Metrics}

We assess the classification accuracy $\mathcal{A}_b$ at each incremental step, with particular emphasis on the final stage's accuracy $\mathcal{A}_B$ as well as the overall average accuracy $\bar{\mathcal{A}}$ across all incremental stages \cite{RN414, RN466, RN467}.

\subsection{Comparison with Baselines}

Comparison results between ours and the baseline methods on different benchmark datasets are illustrated as follows. \textbf{For fairness, all the baseline methods share the same backbone as ours}: ResNet \cite{RN594} as the backbone for image encoding and DGCNN \cite{RN593} (Dynamic Graph CNN) for point encoding; and MLPs are used as the adapter. \textbf{Besides, we share our pre-training method and dataset with the baselines to help them build the image-point correspondence.}

\subsubsection{Results on m-MN40-Inc.4}

\begin{table}[t]
    \centering
     \normalsize
\begin{tabular}{ccc}
\hline
Method    & $\mathcal{A}_B$   \rule{0pt}{10pt}         & $\bar{\mathcal{A}}$            \\ \hline
iCaRL \cite{RN416}     & 25.4          & 48.4          \\
WA  \cite{RN447}    & 20.4          & 40.9          \\
PODNet \cite{RN441}    & 29.0          & 51.9          \\
SimpleCIL \cite{RN448} & 36.1          & 50.2          \\ \hline
Ours      & \textbf{50.8} & \textbf{63.4} \\ \hline
\end{tabular}
    \caption{Comparisons on m-MN40-\textit{Inc}.4.}
    \label{table1}
\end{table}

Table \ref{table1} shows that ours outperforms the baselines by a large margin regarding $\mathcal{A}_B$ and $\bar{\mathcal{A}}$ . iCaRL and WA are classical CIL methods, and PODNet and SImpleCIL received attention recently. They are widely verified in the single-modal CIL. However, in the cross-modal CIL, their performance degrades shapely. These methods cannot balance well between keeping the former category knowledge and the image-point correspondence knowledge.

\begin{figure}[b]
    \centering
    \fbox{
    \includegraphics[width=0.95\linewidth]{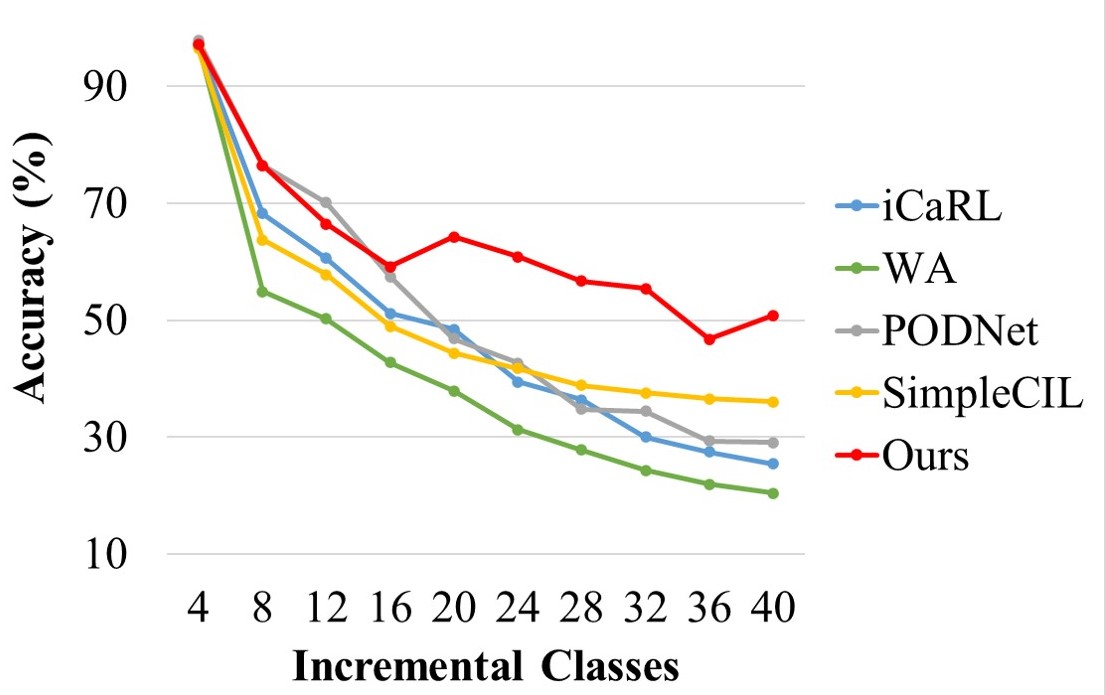}
    }
    \caption{The classification accuracy $\mathcal{A}_b$ at each incremental step with different methods on m-MN40-\textit{Inc}.4}
    \label{fig4}
\end{figure}

The classification accuracy $\mathcal{A}_b$ at each incremental step of different methods on m-MN4-\textit{Inc}.4 is illustrated in Fig. \ref{fig4}. In the prior steps of the incremental learning, the advantage of our methods is not that obvious, even surpassed by PODNet in the 3\textsuperscript{rd} task. While in the later stages, our advantages become increasingly apparent. It proves our method relieves the catastrophic forgetting of the former category knowledge and image-point mappings.

\subsubsection{Results on m-MN40-Inc.8}

Different from m-MN40-\textit{Inc}.4, the experiments on m-MN40-\textit{Inc}.8 aim to discuss the learning ability of the short continual range. Table \ref{table2} shows our method performs better than most baseline methods. However, the advantage is less apparent than that on m-MN40-\textit{Inc}.4. PODNet works better than us regarding $\bar{\mathcal{A}}$, while ours outperforms PODNet on  $\mathcal{A}_B$.

\begin{table}
    \centering
    \normalsize
\begin{tabular}{ccc}
\hline
Method    & $\mathcal{A}_B$   \rule{0pt}{10pt}         & $\bar{\mathcal{A}}$            \\ \hline
iCaRL \cite{RN416}     & 32.2          & 53.4          \\
WA  \cite{RN447}    & 31.9          & 52.3         \\
PODNet \cite{RN441}    & 54.5          & \textbf{68.8}          \\
SimpleCIL \cite{RN448} & 46.0          & 56.1          \\ \hline
Ours      & \textbf{59.4} & 66.8 \\ \hline
\end{tabular}
    \caption{Comparisons on m-MN40-\textit{Inc}.8.}
    \label{table2}
\end{table}

\begin{figure}
    \centering
    \fbox{
    \includegraphics[width=0.95\linewidth]{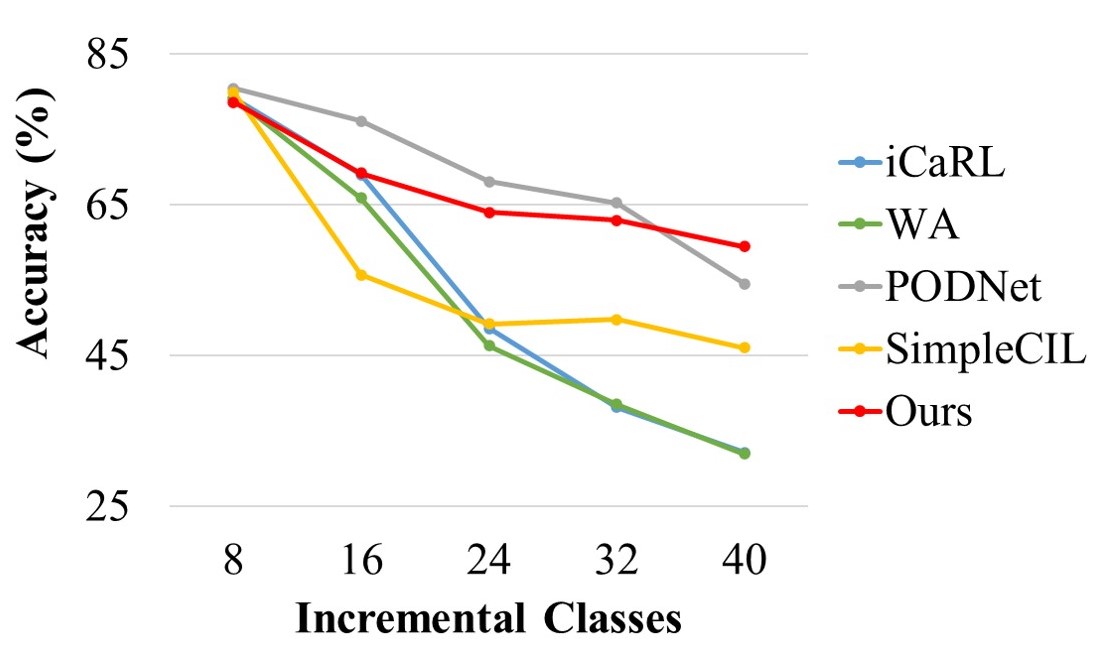}
    }
    \caption{The classification accuracy $\mathcal{A}_b$ at each incremental step with different methods on m-MN40-\textit{Inc}.8}
    \label{fig5}
\end{figure}

We explore deeper experimental phenomena through Fig. \ref{fig5}. It is obvious that even PODNet performs better than us in the prior incremental stages. Our method's forgetting curve has started slowing down, showing better potentiality in preventing model forgetting. It proves that, compared to short-range CIL comparisons, our method works better in long-range ones.

\subsubsection{Results on m-SN55-Inc.6}

\begin{table}
    \centering
    \normalsize
\begin{tabular}{ccc}
\hline
Method    & $\mathcal{A}_B$   \rule{0pt}{10pt}         & $\bar{\mathcal{A}}$            \\ \hline
iCaRL \cite{RN416}     & 38.5          & 57.9          \\
WA  \cite{RN447}    & 24.5          & 43.1          \\
PODNet \cite{RN441}    & 31.1          & 55.2          \\
SimpleCIL \cite{RN448} & 27.3          & 48.9         \\ \hline
Ours      & \textbf{41.9} & \textbf{61.8} \\ \hline
\end{tabular}
    \caption{Comparisons on m-SN55-\textit{Inc}.6.}
    \label{table3}
\end{table}

\begin{figure}
    \centering
    \fbox{
    \includegraphics[width=0.95\linewidth]{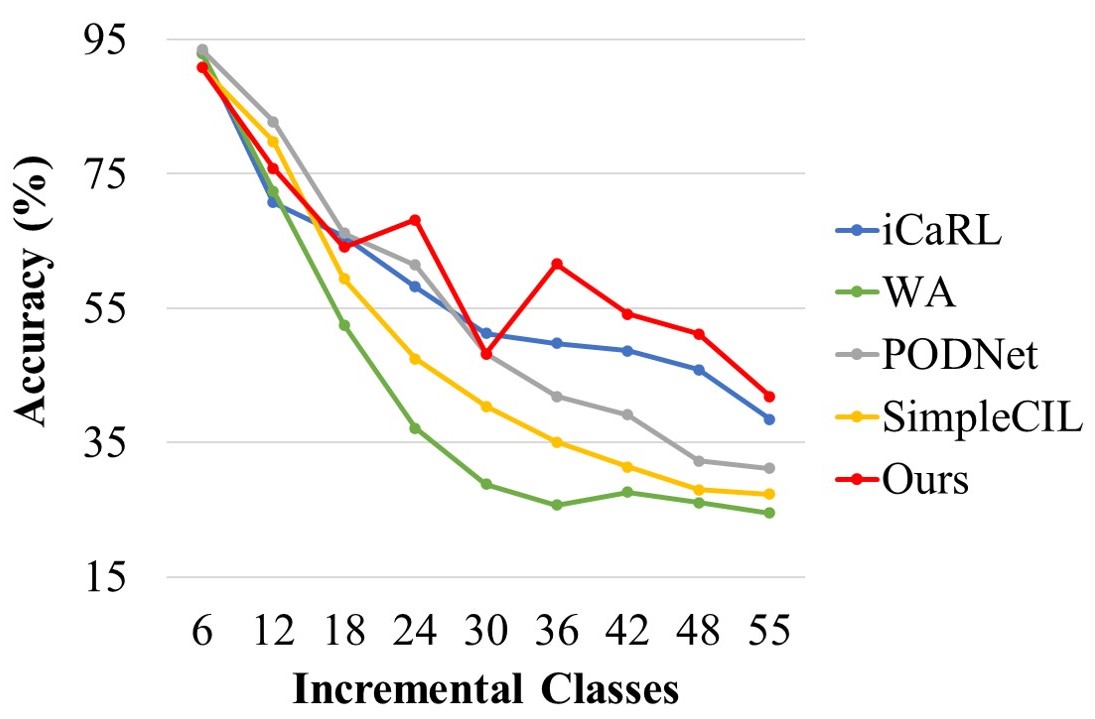}
    }
    \caption{The classification accuracy $\mathcal{A}_b$ at each incremental step with different methods on m-SN55-\textit{Inc}.6}
    \label{fig6}
\end{figure}

Table \ref{table3} compares different methods on m-SN55-\textit{Inc}.6, and our method achieves state-of-the-art results. Fig. \ref{fig6} reports the classification accuracy in different incremental stages. The accuracy of our method degrades sharply in the 5\textsuperscript{th} task. Our method focuses on remembering and adjusting the class prototypes along the incremental stage. However, we only use a single adapter to recalculate the class prototypes for fair comparisons. For the ShapeNet55, which contains lots of objects with different categories but similar geometry characteristics, it is easy to confuse the class prototypes, leading to a sharp degradation in the classification accuracy $\mathcal{A}_b$.

\subsection{Qualitative Analysis}

\begin{figure}
    \centering
    \includegraphics[width=0.7\linewidth]{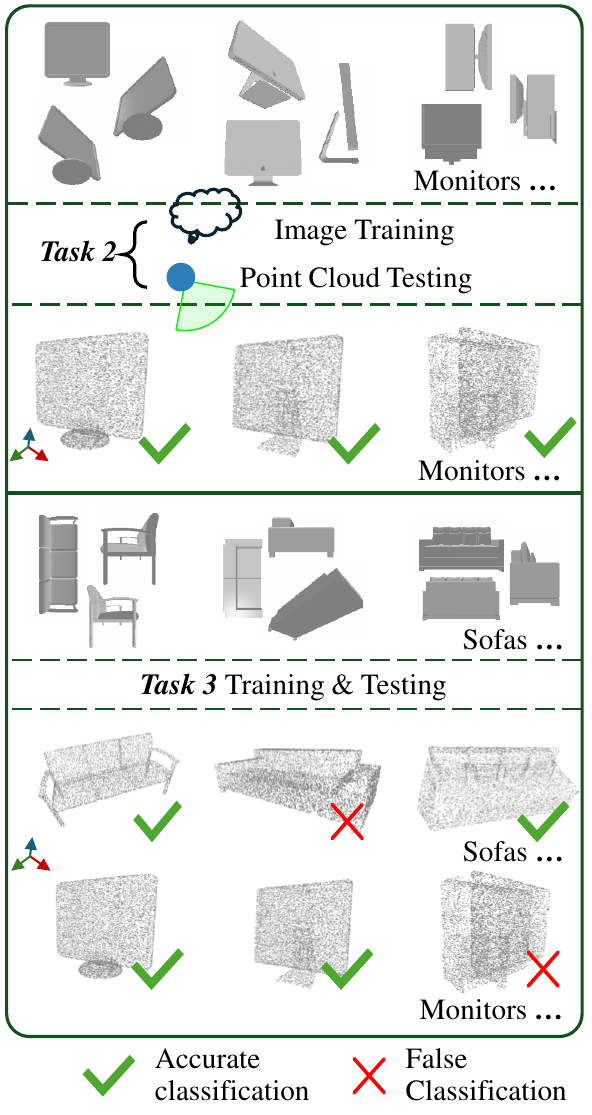}
    \caption{Visualizations of our method's continual learning results on m-MN40-\textit{Inc}.4. Take monitor and sofa classifications in tasks 2 and 3 as examples (omit other categories of objects): in task 2, learn to classify the image-based monitors, testing point-cloud-based monitors; in task 3, learn to classify the image-based sofas, testing point-cloud-based monitors and sofas. Our method misclassifies some monitors and sofas.}
    \label{fig7}
\end{figure}

Fig. \ref{fig7} visualizes some experimental results of our method on MN40-\textit{Inc}.4. Our method learns to classify image-based monitor in task 2, transferring the category knowledge from the images to the point clouds and maintaining high accuracy while testing point cloud classifications. Our method still maintains high accuracy for monitor and sofa classification in task 3. But some monitors and sofas are misclassified.

As discussed above, the biggest challenge of the IP-CIL is to relieve the model's cross-modal catastrophic forgetting, not only the former category knowledge but also the generalizable correspondence between images and point clouds. Our CMIP-CIL method focuses on addressing this challenge, but it is still impossible to completely avoid knowledge forgetting. Our method forgets some previously learned monitor category knowledge learned in task 3, thus misclassifying some monitor samples. Besides, the generalization ability of image-point correspondence does not cover all the sofas, leading to some sofa misclassifications.

\subsection{Ablation Study}

We conduct ablation studies to verify the effectiveness of some essential designs in our method: image rendering with point random masking (RRM) in section \textit{IV-B} and the regularizations of prototype calculation in section \textit{IV-D}. We also verify the effectiveness of the temperature coefficient in Eq. (\ref{eq1})

\subsubsection{Effectiveness of RRM for Image-Point Correspondence Establishment}

RRM is the core of our method, empowering the model with the generalizations of image-point correspondence. To prove RRM's effectiveness, we remove it and only pre-train the model on the multi-modal dataset without masking. Table \ref{table4} illustrates the experimental results: the accuracy of tasks 1, 2 ($\mathcal{A}_1$, $\mathcal{A}_2$) decreases significantly (the accuracy stays so low in the latter tasks without our RRM; thus, we do not report them). 

\begin{table}
    \centering
    \normalsize
\begin{tabular}{cccccc}
\hline
RRM & \begin{tabular}[c]{@{}c@{}}Regularization\\ Item\end{tabular} & $\mathcal{A}_1$ & $\mathcal{A}_2$  & $\mathcal{A}_B$  & $\bar{\mathcal{A}}$  \\ \hline
w/o & w/                                                            & 34.2 & 15.8  & -  & -  \\
w/  & w/o                                                           & 96.8 &  \textbf{77.6}  & 46.5  & 62.1  \\ \hline
w/  & w/                                                            & \textbf{97.2} & 76.5 & \textbf{50.8} & \textbf{63.4} \\ \hline
\end{tabular}
    \caption{Experimental results on m-MN40-\textit{Inc}.4 considering RRM and regularizations in Prototype Calculation.}
    \label{table4}
\end{table}

This ablation weakens the learnings of the image-point correspondence. Thus, our method cannot transfer the category knowledge from images to the point cloud well. This ablation degrades the generalization ability of cross-modal mapping; thus, the performance degradation is more obvious on datasets that have not been seen in pre-training ($\mathcal{A}_2$ in task 2 as an example).

\subsubsection{Effectiveness of the Regularizations in Prototype Calculation}

The regularization item promotes the representations of objects with the same class category to approach the class prototype. Thus, our model can memorize and adjust the prototypes along the incremental stage, relieving the model's forgetting of previous category knowledge. We remove the regularization in the prototype calculation to verify this item's effectiveness. Table \ref{table4} illustrates the experimental results: In the initial stages, the advantage of the regularization term is not significant, as the performance with and without the regularization term ($\mathcal{A}_1$, $\mathcal{A}_2$) shows little difference. However, in later stages, the advantage of the regularization term gradually becomes more apparent, with $\mathcal{A}_B$  and $\bar{\mathcal{A}}$ showing significant lead.

Without the regularization item, even for the objects with the same category, the representations may be diverse. Thus, it poses a big challenge for the model to remember the characteristics of each observed object. Thus, the regularization item in the prototype calculation is a beneficial design for this cross-modal continual learning task.

\subsubsection{Effectiveness of the Temperature Coefficient $\tau$ in Eq. (\ref{eq1})}

\begin{table}
    \centering
    \normalsize
\begin{tabular}{cll}
\hline
\begin{tabular}[c]{@{}c@{}}$\tau$\\ in Eq. (\ref{eq1})\end{tabular} & $\mathcal{A}_B$ & $\bar{\mathcal{A}}$ \\ \hline
0.01                                                                                                 & \textbf{51.9} &  61.3 \\

0.04                                                                                                  & 47.2 &  62.1 \\ \hline
0.02                                                                                                  & 50.8 & \textbf{63.4} \\ \hline
\end{tabular}
    \caption{Experimental results on m-MN40-\textit{Inc}.4 considering different temperature coefficients in Eq. (\ref{eq1}).}
    \label{table5}
\end{table}

The temperature coefficient ($\tau$) in NT-Xent loss scales the logits in the softmax function, controlling similarity sharpness. Lower $\tau$ emphasizes hard negatives, enhancing clustering but risking overfitting. Higher $\tau$ smoothens the distribution, promoting broader exploration and better generalization, making $\tau$ essential in image-point contrastive learning. Table \ref{table5} shows the experimental results considering different $\tau$.

Setting $\tau$ = 0.02 achieves the best results regarding $\mathcal{A}_B$ and $\bar{\mathcal{A}}$. $\tau$ = 0.01 makes the model overemphasize hard negatives, leading to a worse generalization of image-point correspondence. $\tau$ = 0.04 blurs sample distinctions and weakens the feature discriminations. Thus, $\tau$ = 0.02 is the best choice in the contrastive masked image-point pre-training.

\section{CONCLUSIONS}

In this paper, we address the challenging task of image-point class incremental learning, aiming to continually learn category knowledge from 2D images and enhance robots' perceptual capabilities in dynamic environments. Our method effectively mitigates cross-modal catastrophic forgetting, enabling the model to retain and generalize knowledge across previously seen categories while seamlessly learning new ones. Through comprehensive experiments on benchmark datasets, our method demonstrates state-of-the-art performance, significantly outperforming baseline approaches. Future work will focus on further enhancing the model's adaptability to more complex cross-modal scenarios and exploring its applications in real-world robotic systems.





\section*{ACKNOWLEDGMENT}

This work was supported by the National Natural Science Foundation of China (Grant No. 62403491).


\bibliographystyle{ieeetr}

\end{document}